\documentclass{article}
\usepackage{ijcai19}

\usepackage{times}
\usepackage{soul}
\usepackage{url}
\usepackage[hidelinks]{hyperref}
\usepackage[utf8]{inputenc}
\usepackage[small]{caption}
\usepackage{subfigure}
\usepackage{graphicx}
\usepackage{amsmath}
\usepackage{amssymb}
\usepackage{booktabs}
\usepackage{algorithm}
\usepackage{algorithmic}
\usepackage{balance}
\usepackage{enumitem}
\title{Representation Learning-Assisted Click-Through Rate Prediction}

\author{
    Wentao Ouyang, Xiuwu Zhang, Shukui Ren, Chao Qi, Zhaojie Liu \textnormal{and} Yanlong Du
    \affiliations
    Intelligent Marketing Platform, Alibaba Group
    \emails
    \{maiwei.oywt, xiuwu.zxw, shukui.rsk, qichao.qc, zhaojie.lzj, yanlong.dyl\}@alibaba-inc.com
}

\begin{document}

\maketitle

\begin{abstract}
Click-through rate (CTR) prediction is a critical task in online advertising systems. Most existing methods mainly model the feature-CTR relationship and suffer from the data sparsity issue. In this paper, we propose DeepMCP, which models other types of relationships in order to learn more informative and statistically reliable feature representations, and in consequence to improve the performance of CTR prediction. In particular, DeepMCP contains three parts: a matching subnet, a correlation subnet and a prediction subnet. These subnets model the user-ad, ad-ad and feature-CTR relationship respectively. When these subnets are jointly optimized under the supervision of the target labels, the learned feature representations have both good prediction powers and good representation abilities. Experiments on two large-scale datasets demonstrate that DeepMCP outperforms several state-of-the-art models for CTR prediction.
\end{abstract}

\section{Introduction}
Click-through rate (CTR) prediction is to predict the probability that a user will click on an item. It plays an important role in online advertising systems. For example, the ad ranking strategy generally depends on CTR $\times$ bid, where bid is the benefit the system receives if an ad is clicked by a user. Moreover, according to the common cost-per-click charging model, advertisers are only charged once their ads are clicked by users. Therefore, in order to maximize the revenue and to maintain a desirable user experience, it is crucial to estimate the CTR of ads accurately.

CTR prediction has attracted lots of attention from both academia and industry \cite{he2014practical,shan2016deep,guo2017deepfm}.
For example, the Logistic Regression (LR) model \cite{richardson2007predicting} considers linear feature importance and models the predicted CTR as
$
\hat{y} = \sigma (w_0 + \sum_i w_i x_i),
$
where $\sigma(\cdot)$ is the sigmoid function, $x_i$ is the $i$th feature and $w_0, w_i$ are model weights.
The Factorization Machine (FM) \cite{rendle2010factorization} is proposed to further model pairwise feature interactions. It models the predicted CTR as
$
\hat{y} = \sigma (w_0 + \sum_i w_i x_i + \sum_i \sum_j \mathbf{v}_i^T \mathbf{v}_j x_i x_j),
$
where $\mathbf{v}_i$ is the latent embedding vector of the $i$th feature.
In recent years, Deep Neural Networks (DNNs) \cite{lecun2015deep} are exploited for CTR prediction and item recommendation in order to automatically learn feature representations and high-order feature interactions \cite{van2013deep,zhang2016deep,qu2016product,covington2016deep}. To take advantage of both shallow and deep models, hybrid models are also proposed. For example, Wide\&Deep \cite{cheng2016wide} combines LR and DNN, in order to improve both the memorization and generalization abilities of the model.
DeepFM \cite{guo2017deepfm} combines FM and DNN, which further improves the model ability of learning feature interactions.
Neural Factorization Machine \cite{he2017neural} combines the linearity of FM and the non-linearity of DNN.

\begin{figure}[!t]
\centering
\subfigure[]{\includegraphics[width=0.23\textwidth]{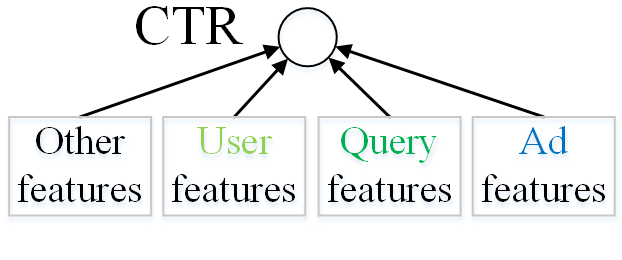}} \hskip 3pt
\subfigure[]{\includegraphics[width=0.24\textwidth]{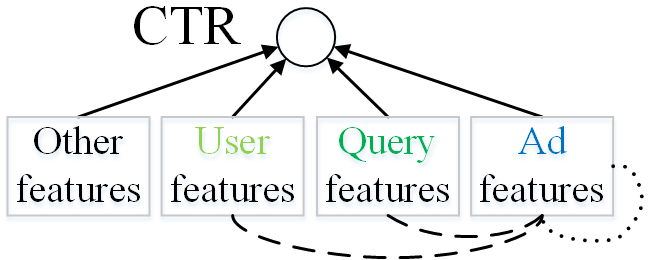}}
\vskip -8pt
\caption{(a) Classical CTR prediction methods model the feature-CTR relationship. (b) DeepMCP further models feature-feature relationships, such as the user-ad relationship (dashed curve) and the ad-ad relationship (dotted curve).}
\vskip -10pt
\label{illus}
\end{figure}

Nevertheless, these models only consider the \textbf{feature-CTR} relationship. In contrast, the DeepMCP model proposed in this paper additionally considers \textbf{feature-feature} relationships, such as the \textbf{user-ad} relationship and the \textbf{ad-ad} relationship. We illustrate their difference in Figure \ref{illus}. Note that the feature interaction in FM still models the feature-CTR relationship. It can be considered as two\_features-CTR, because it models how the feature interaction $\mathbf{v}_i^T \mathbf{v}_j x_i x_j$ relates to the CTR $\hat{y}$, but does not model whether the two feature representations $\mathbf{v}_i$ and $\mathbf{v}_j$ should be similar to each other.

In particular, our proposed DeepMCP model contains three parts: a matching subnet, a correlation subnet and a prediction subnet. They share the same embedding matrix.
The matching subnet models the user-ad relationship (i.e., whether an ad matches a user's interest) and aims to learn useful user and ad representations.
The correlation subnet models the ad-ad relationship (i.e., which ads are within a time window in a user's click sequence) and aims to learn useful ad representations.
The prediction subnet models the feature-CTR relationship and aims to predict the CTR given all the features.
When these subnets are jointly optimized under the supervision of the target labels, the feature representations are learned in such a way that they have both good prediction powers and good representation abilities. Moreover, as the same feature appears in different subnets in different ways, the learned representations are more statistically reliable.

In summary, the main contributions of this paper are
\begin{itemize} 
\item We propose a new model DeepMCP for CTR prediction. Unlike classical CTR prediction models that mainly consider the feature-CTR relationship, DeepMCP further considers user-ad and ad-ad relationships.
\item We conduct extensive experiments on two large-scale datasets to compare the performance of DeepMCP with several state-of-the-art models. We make the implementation code of DeepMCP publicly available\footnote{https://github.com/oywtece/deepmcp}.
\end{itemize}

\section{Deep Matching, Correlation and Prediction (DeepMCP) Model} \label{sec_deepmcp}
In this section, we present the DeepMCP model in detail.

\begin{table}[!t]
\setlength{\tabcolsep}{3pt}
\renewcommand{\arraystretch}{1.2}
\centering
\begin{tabular}{|c|c|c|c|c|}
\hline
\textbf{Label} & \textbf{User ID} & \textbf{User Age} & \textbf{Ad Title} \\
\hline
1 & 2135147 & 24 & Beijing flower delivery \\
\hline
0 & 3467291 & 31 & Nike shoes, sporting shoes \\
\hline
0 & 1739086 & 45 & Female clothing and jeans \\
\hline
\end{tabular}
\vskip -4pt
\caption{Each row is an instance for CTR prediction. The first column is the label (1 - clicked, 0 - unclicked). Each of the other columns is a field. The instantiation of a field is a feature.}
\label{tab_ft}
\vskip -4pt
\end{table}

\begin{figure}[!t]
\centering
\subfigure[DeepMCP - Training]{\includegraphics[width=0.45\textwidth]{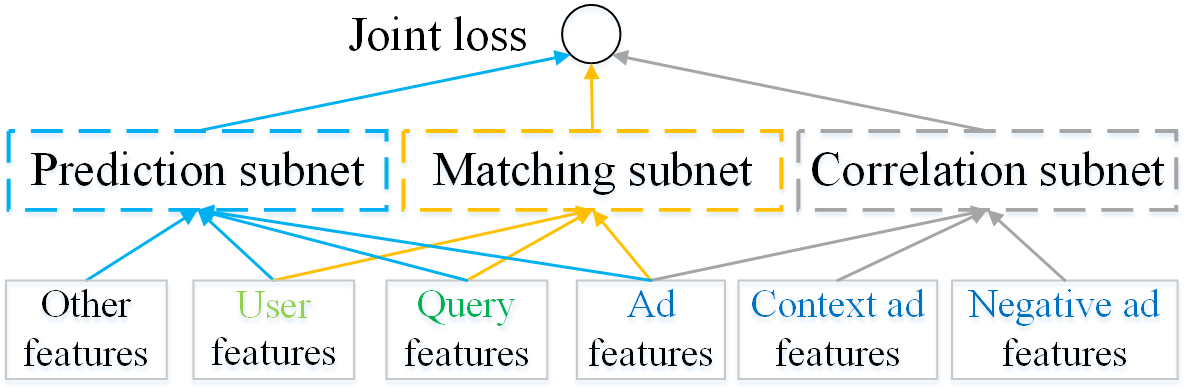}} \vskip -5pt
\subfigure[DeepMCP - Testing]{\includegraphics[width=0.45\textwidth]{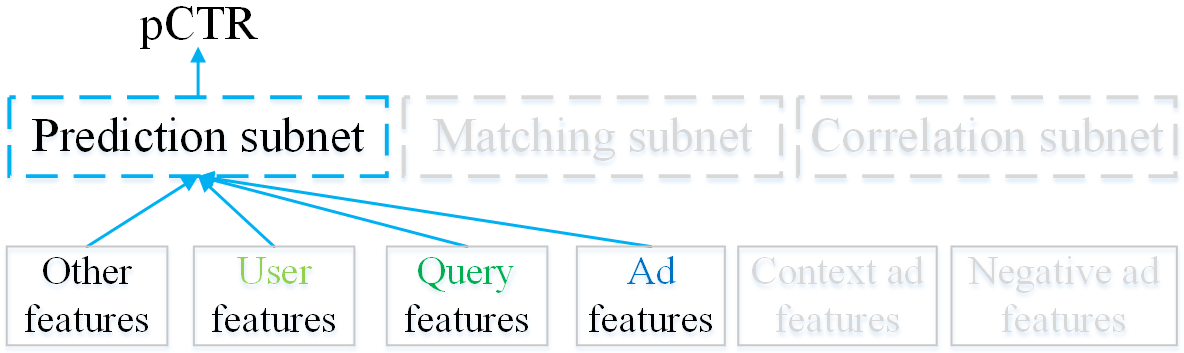}}
\vskip -4pt
\caption{Sketch view of the DeepMCP model: (a) Training, (b) Testing. All the subnets are active during training, but only the prediction subnet is active during testing. pCTR is predicted CTR.}
\vskip -6pt
\label{model_sketch}
\end{figure}

\subsection{Model Overview} \label{sec_overview}
The task of CTR prediction in online advertising is to estimate the probability of a user clicking on a specific ad.
Table \ref{tab_ft} shows some example instances. Each instance can be described by multiple \emph{fields} such as user information (user ID, city, etc.) and ad information (creative ID, title, etc.). The instantiation of a field is a \emph{feature}.

Unlike most existing CTR prediction models that mainly consider the feature-CTR relationship, our proposed DeepMCP model additionally considers the user-ad and ad-ad relationships.
DeepMCP contains three parts: a matching subnet, a correlation subnet and a prediction subnet (cf. Figure \ref{model_sketch}(a)).
When these subnets are jointly optimized under the supervision of the target labels, the learned feature representations have both good prediction powers and good representation abilities.
Another property of DeepMCP is that although all the subnets are active during training, only the prediction subnet is active during testing (cf. Figure \ref{model_sketch}(b)). This makes the testing phase rather simple and efficient.

\begin{figure}[!t]
\centering
\includegraphics[width=0.42\textwidth]{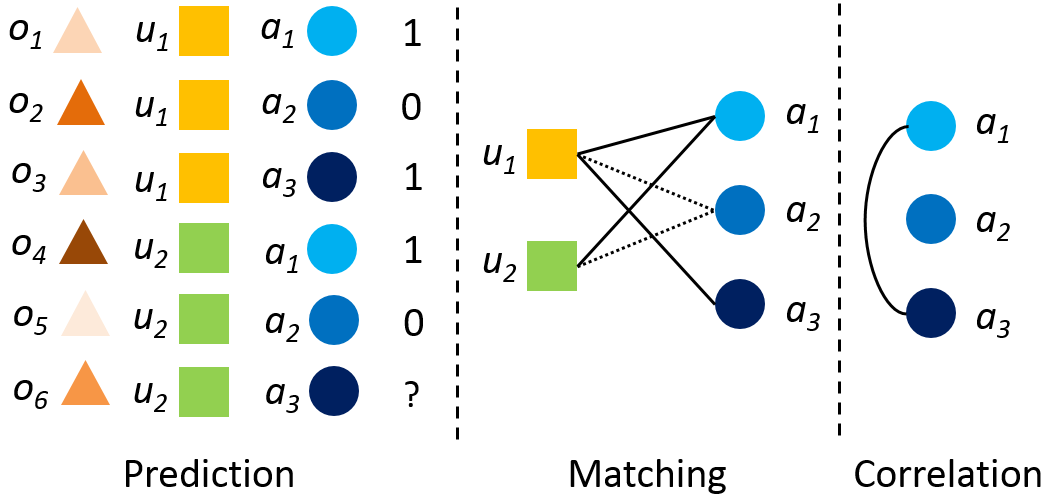}
\vskip -4pt
\caption{Motivating example ($u$ - user features, $a$ - ad features, $o$ - other features). Please refer to \S\ref{sec_overview} for more detail.}
\vskip -6pt
\label{example}
\end{figure}

We segregate the features into four groups: \emph{user} (e.g., user ID, age), \emph{query} (e.g., query, query category), \emph{ad} (e.g., creative ID, ad title) and \emph{other} features (e.g., hour of day, day of week). Each subnet uses a different set of features. In particular, the prediction subnet uses all the four groups of features, the matching subnet uses the user, query and ad features, and the correlation subnet uses only the ad features.
All the subnets share \emph{the same} embedding matrix.

\begin{figure*}[!t]
\centering
\includegraphics[width=0.98\textwidth]{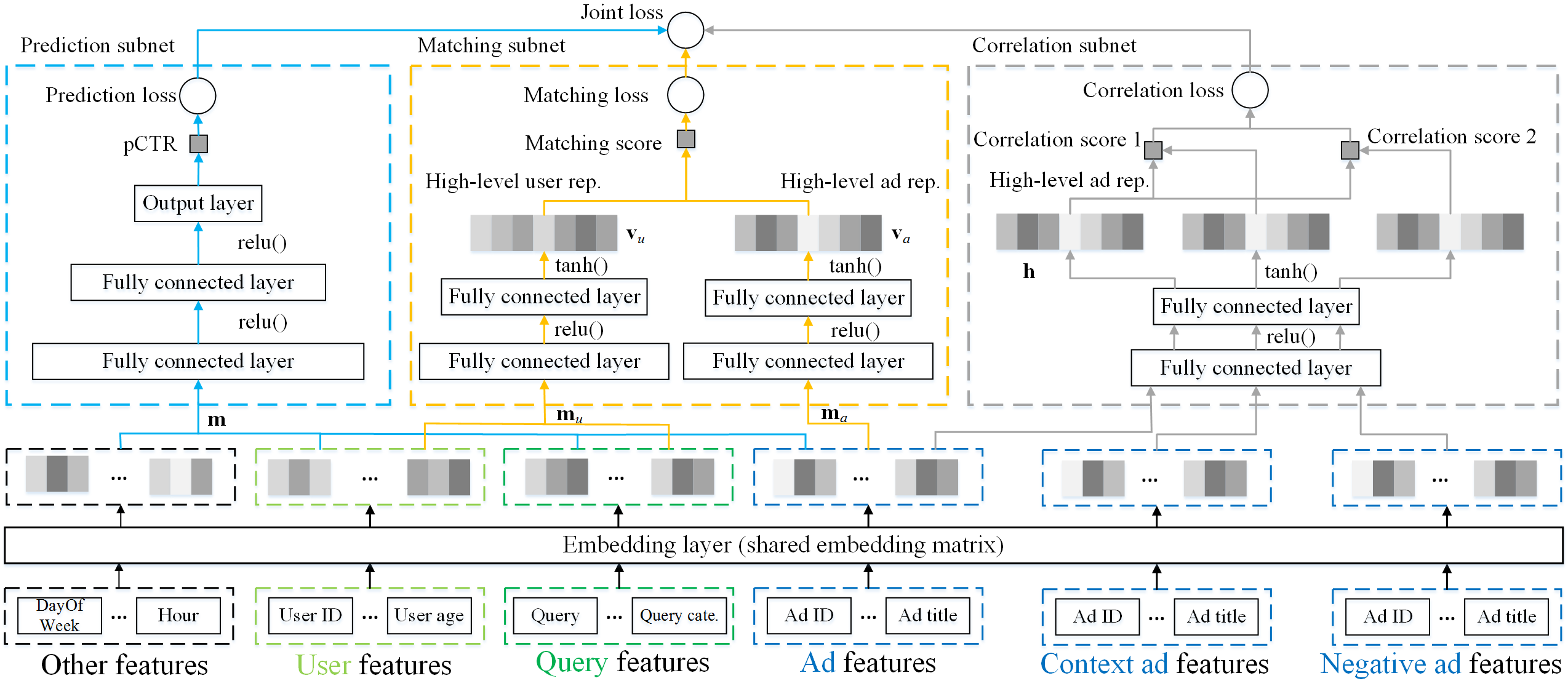}
\vskip -4pt
\caption{Detailed view of the DeepMCP model. The prediction, matching and correlation subnets share \emph{the same} embedding matrix.}
\vskip -6pt
\label{model_detail}
\end{figure*}

\subsubsection{Motivating Example}
Before we present the details of DeepMCP, we first illustrate the rationale of DeepMCP through a motivating example in Figure \ref{example}. For simplicity, we only show user features $u$, ad features $a$  and other features $o$.

Because the feature embeddings (i.e., representations) are randomly initialized, when we consider the prediction task only, it is likely that the learned representation of user $u_1$ and that of user $u_2$ are largely different. This is because the prediction task does not model the relationship between features.
As a consequence, it is hard to accurately estimate the pCTR of user $u_2$ on ad $a_3$.
If we further consider the matching task which models the user-ad relationship and the correlation task which models the ad-ad relationship, the learned representation of user $u_2$ should be similar to that of user $u_1$ and the representation of ad $a_3$ should be similar to that of ad $a_1$. The pCTR of user $u_2$ on ad $a_3$ would be similar to the pCTR of $u_1$ on $a_3$ (as well as the pCTR of $u_1$ on $a_1$). As a consequence, the target pCTR is more likely to be accurate.

\subsection{Prediction Subnet}
The prediction subnet presented here is a typical DNN model. It models the feature-CTR relationship (where explicit or implicit feature interactions are modeled). It aims to predict the CTR given all the features, supervised by the target labels.
Nevertheless, the DeepMCP model is flexible that the prediction subnet can be replaced by any other CTR prediction model, such as Wide\&Deep \cite{cheng2016wide} and DeepFM \cite{guo2017deepfm}.

First, a feature $x_i \in \mathbb{R}$ (e.g., a user ID) goes through an embedding layer and is mapped to its embedding vector $\mathbf{e}_i \in \mathbb{R}^K$, where $K$ is the vector dimension and $\mathbf{e}_i$ is to be learned. The collection of all the feature embeddings is an embedding matrix $\mathbf{E} \in \mathbb{R}^{N\times K}$, where $N$ is the number of unique features. For multivalent categorical features such as the bi-grams in the ad title, we first map each bi-gram to an embedding vector and then perform sum pooling to generate the aggregated embedding vector of the ad title.

We then concatenate the embedding vectors from all the features as a long vector $\mathbf{m}$.
The vector $\mathbf{m}$ then goes through several fully connected (FC) layers with the ReLU activation function ($\mathrm{ReLU}(x) = \max(0, x)$), in order to exploit high-order nonlinear feature interactions \cite{he2017neural}. Nair and Hinton [\citeyear{nair2010rectified}] show that ReLU has significant benefits over sigmoid and tanh activation functions in terms of the convergence rate and the quality of obtained results.

Finally, the output $\mathbf{z}$ of the last FC layer goes through a sigmoid function to generate the predicted CTR as
\[
\hat{y} = \frac{1}{1+\exp[- (\mathbf{w}^T \mathbf{z} + b)]},
\]
where $\mathbf{w}$ and $b$ are model parameters to be learned.
To avoid model overfitting, we apply dropout \cite{srivastava2014dropout} after each FC layer.
Dropout prevents feature co-adaptation by setting to zero a portion of hidden units during parameter learning.

All the model parameters are learned by minimizing the average logistic loss on a training set as
\begin{equation} \label{loss_p}
\mathrm{loss}_p = - \frac{1}{|\mathbb{Y}|}\sum_{y \in \mathbb{Y}} [y \log \hat{y} + (1 - y) \log (1 - \hat{y})],
\end{equation}
where $y \in\{0,1\}$ is the true label of the target ad corresponding to $\hat{y}$ and $\mathbb{Y}$ is the collection of labels.

\subsection{Matching Subnet}
The matching subnet models the user-ad relationship (i.e., whether an ad matches a user's interest) and aims to learn useful user and ad representations. It is inspired by semantic matching models for web search \cite{huang2013learning}.

In classical matrix factorization for recommendation \cite{koren2009matrix}, the rating score is approximated as the inner product of the latent vectors of the user ID and the item ID.
In our problem, instead of directly matching the user ID and the ad ID, we perform matching at a \emph{higher level}, incorporating all the features related to the user and the ad.
When a user clicks an ad, we assume that the clicked ad is relevant, at least partially, to the user's need (given the query submitted by the user, if any).
In consequence, we would like the representation of the user features (and the query features) and the representation of the ad features to match well.

In particular, the matching subnet contains two parts: ``user part'' and ``ad part''.
The input to the ``user part'' is the user features (e.g., user ID, age) and query features (e.g., query, query category).
As in the prediction subnet, a feature $x_i \in \mathbb{R}$ first goes through an embedding layer and is mapped to its embedding vector $\mathbf{e}_i \in \mathbb{R}^K$.
We then concatenate all the feature embeddings as a long vector $\mathbf{m}_u \in \mathbb{R}^{N_u}$ ($N_u$ is the vector dimension).
The vector $\mathbf{m}_u$ then goes through several FC layers in order to learn more abstractive, high-level representations. We use $\tanh$ (rather than ReLU) as the activation function of the last FC layer, which is defined as
$
\tanh(x) = \frac{1-\exp(-2x)}{1+\exp(-2x)}.
$
We will explain the reason later.
The output of the ``user part'' is a high-level user representation vector $\mathbf{v}_u \in \mathbb{R}^M$ ($M$ is the vector dimension).

The input to the ``ad part'' is the ad features (e.g., creative ID, ad title). Similarly, we first map each ad feature to its embedding vector and then concatenate them as a long embedding vector $\mathbf{m}_a \in \mathbb{R}^{N_a}$ ($N_a$ is the vector dimension).
The vector $\mathbf{m}_a$ then goes through several FC layers and results in a high-level ad representation vector $\mathbf{v}_a \in \mathbb{R}^M$.
Note that, the inputs to the ``user'' and ``ad'' parts usually have different sizes, i.e., $N_u \neq N_a$ (because the number of user features and the number of ad features may not necessarily be the same). However, after the matching subnet, $\mathbf{v}_u$ and $\mathbf{v}_a$ have the same size $M$. In other words, we project two different sets of features into a common low-dimensional space.

We then compute the matching score $s$ as
\[
s(\mathbf{v}_u, \mathbf{v}_a) = \frac{1}{1 + \exp(- \mathbf{v}_u^T \mathbf{v}_a)}.
\]
We do not use ReLU as the activation function of the last FC layer because the output after ReLU will contain lots of zeros, which makes $\mathbf{v}_u^T \mathbf{v}_a \rightarrow 0$.
There are at least two choices to model the matching score: point-wise and pair-wise \cite{liu2009learning}. In a point-wise model, we could model $s(\mathbf{v}_u, \mathbf{v}_a) \rightarrow 1$ if user $u$ clicks ad $a$ and model $s(\mathbf{v}_u, \mathbf{v}_a) \rightarrow 0$ otherwise. In a pair-wise model, we could model $s(\mathbf{v}_u, \mathbf{v}_{a_i}) > s(\mathbf{v}_u, \mathbf{v}_{a_j}) + \delta$ where $\delta >0$ is a margin, if user $u$ clicks ad $a_i$ but not ad $a_j$.

We choose the point-wise model because it can directly reuse the training dataset for the prediction subnet.
Formally, we minimize the following loss for the matching subnet
\begin{align}
\mathrm{loss}_m &= - \frac{1}{|\mathbb{Y}|}\sum_{y \in \mathbb{Y}} \big[y(u,a) \log s(\mathbf{v}_u, \mathbf{v}_a) \nonumber \\
&+ (1 - y(u,a)) \log (1 - s(\mathbf{v}_u, \mathbf{v}_a)) \big], \label{loss_m}
\end{align}
where $y(u,a) = 1$ if user $u$ clicks ad $a$ and it is $0$ otherwise.

\subsection{Correlation Subnet}
The correlation subnet models the ad-ad relationship (i.e., which ads are within a time window in a user's click sequence) and aims to learn useful ad representations.
The skip-gram model is proposed in \cite{mikolov2013distributed} to learn useful representations of words in a sequence, where words within a context window have certain correlation. It has been widely applied in many tasks to learn useful low-dimensional representations \cite{zhao2018learning,zhou2018deep}.

In our problem, we apply the skip-gram model to learn useful ad representations, because the clicked ads of a user also form a sequence with certain correlation over time. Formally, given a sequence of ads $\{a_1, a_2, \cdots, a_L\}$ clicked by a user, we would like to maximize the average log likelihood as
\[
ll = \frac{1}{L} \sum_{i=1}^L \sum_{-C \leq j \leq C}^{1\leq i+j \leq L, j \neq 0} \log p(a_{i+j} | a_i),
\]
where $L$ is the number of ads in the sequence and $C$ is a context window size.

The probability $p(a_{i+j} | a_i)$ can be defined in different ways such as softmax, hierarchical softmax and negative sampling \cite{mikolov2013distributed}. We choose the negative sampling technique due to its efficiency. $p(a_{i+j} | a_i)$ is then defined as
\[
p(a_{i+j} | a_i) = \sigma(\mathbf{h}_{a_{i+j}}^T \mathbf{h}_{a_i}) \prod_{q=1}^Q  \sigma(-\mathbf{h}_{a_q}^T \mathbf{h}_{a_i}),
\]
where $Q$ is the number of sampled negative ads and
$
\sigma(\mathbf{h}_{a_{i+j}}^T \mathbf{h}_{a_i}) = \frac{1}{1 + \exp(- \mathbf{h}_{a_{i+j}}^T \mathbf{h}_{a_i})}.
$
$\mathbf{h}_{a_i}$ is a high-level representation vector that involves all the features related to ad $a_i$ and that goes through several FC layers (cf. Figure \ref{model_detail}).

The loss function of the correlation subnet is then given by minimizing the negative average log likelihood as
\begin{align}
\mathrm{loss}_c & = \frac{1}{L} \sum_{i=1}^L \sum_{-C \leq j \leq C}^{1\leq i+j \leq L, j \neq 0} \Big[- \log \left[\sigma(\mathbf{h}_{a_{i+j}}^T \mathbf{h}_{a_i}) \right] \nonumber \\
& - \sum_{q=1}^Q \log \left[\sigma(-\mathbf{h}_{a_q}^T \mathbf{h}_{a_i}) \right] \Big]. \label{loss_c}
\end{align}

\subsection{Offline Training Procedure}
The final joint loss function of DeepMCP is given by
\begin{equation} \label{loss}
\mathrm{loss} = \mathrm{loss}_p + \alpha \mathrm{loss}_m + \beta \mathrm{loss}_c,
\end{equation}
where $\mathrm{loss}_p$ is the prediction loss in Eq. (\ref{loss_p}), $\mathrm{loss}_m$ is the matching loss in Eq. (\ref{loss_m}), $\mathrm{loss}_c$ is the correlation loss in Eq. (\ref{loss_c}), and $\alpha$ and $\beta$ are tunable hyperparameters for balancing the importance of different subnets.

The DeepMCP model is trained by minimizing the joint loss function on a training dataset.
Since our aim is to maximize the CTR prediction performance, we evaluate the model on a separate validation dataset and record the validation AUC (an evaluation metric, which will be explained in \S\ref{sec_metric}) during the training procedure. The optimal model parameters are obtained at the highest validation AUC.

\subsection{Online Procedure}
As we have illustrated in Figure \ref{model_sketch}(b), in the online testing phase, the DeepMCP model only needs to compute the predicted CTR (pCTR). Therefore, only the features from the target ad are needed and only the prediction subnet is active. This makes the online phase of DeepMCP rather simple and efficient.

\section{Experiments}
In this section, we conduct experiments on two large-scale datasets to evaluate the performance of DeepMCP and several state-of-the-art methods for CTR prediction.

\subsection{Datasets}
\begin{table}[!t]
\renewcommand{\arraystretch}{1.2}
\centering
\begin{tabular}{|l|l|l|l|l|l|}
\hline
\textbf{Dataset} & \textbf{\# Instances} & \textbf{\# Fields} & \textbf{\# Features} \\
\hline
Avito & 11,211,794 & 27 & 42,301,586 \\
\hline
Company & 60,724,276 & 28 & 29,997,247 \\
\hline
\end{tabular}
\vskip -4pt
\caption{Statistics of experimental large-scale datasets.}
\label{tab_stat}
\end{table}

Table \ref{tab_stat} lists the statistics of two large-scale datasets.

1) \textbf{Avito advertising dataset\footnote{https://www.kaggle.com/c/avito-context-ad-clicks/data}.}
This dataset contains a random sample of ad logs from avito.ru, the largest general classified website in Russia.
We use the ad logs from 2015-04-28 to 2015-05-18 for training, those on 2015-05-19 for validation, and those on 2015-05-20 for testing. In CTR prediction, testing is usually the next-day prediction. The test set contains $2.3\times 10^6$ instances.
The features used include 1) user features such as user ID, IP ID, user agent and user device, 2) query features such as search query, search category and search parameters, 3) ad features such as ad ID, ad title and ad category, and 4) other features such as hour of day and day of week.

2) \textbf{Company advertising dataset.}
This dataset contains a random sample of ad impression and click logs from a commercial advertising system in Alibaba. We use ad logs of 30 consecutive days during Aug.-Sep. 2018 for training, logs of the next day for validation, and logs of the day after the next day for testing.  The test set contains $1.9\times 10^6$ instances.
The features used also include user, query, ad and other features.

\subsection{Methods Compared}
We compare the following methods for CTR prediction.
\begin{enumerate} 
\item \textbf{LR}. Logistic Regression \cite{richardson2007predicting}. It models linear feature importance.
\item \textbf{FM}. Factorization Machine \cite{rendle2010factorization}. It models both first-order feature importance and second-order feature interactions.
\item \textbf{DNN}. Deep Neural Network. It contains an embedding layer, several fully connected layers and an output layer.
\item \textbf{PNN}. The Product-based Neural Network in \cite{qu2016product}. It introduces a production layer between the embedding layer and fully connected layers of DNN.
\item \textbf{Wide\&Deep}. The Wide\&Deep model in \cite{cheng2016wide}. It combines LR (wide part) and DNN (deep part).
\item \textbf{DeepFM}. The DeepFM model in \cite{guo2017deepfm}. It combines FM (wide part) and DNN (deep part).
\item \textbf{DeepCP}. A variant of the DeepMCP model, which contains only the correlation and the prediction subnets. It is equivalent to setting $\alpha = 0$ in Eq. (\ref{loss}).
\item \textbf{DeepMP}. A variant of the DeepMCP model, which contains only the matching and the prediction subnets. It is equivalent to setting $\beta=0$ in Eq. (\ref{loss}).
\item \textbf{DeepMCP}. The DeepMCP model (\S\ref{sec_deepmcp}) which contains the matching, correlation and prediction subnets.
\end{enumerate}

\subsection{Parameter Settings}
We set the embedding dimension of each feature as $K=10$, because the number of distinct features is huge.
We set the number of fully connected layers in neural network-based models as 2, with dimensions 512 and 256. We set the batch size as 128, the context window size as $C=2$ and the number of negative ads as $Q=4$.
The dropout ratio is set to 0.5. All the methods are implemented in Tensorflow and optimized by the Adagrad algorithm \cite{duchi2011adaptive}.

\begin{table}[!t]
\setlength{\tabcolsep}{4pt}
\renewcommand{\arraystretch}{1.3}
\centering
\begin{tabular}{|l|c|c||c|c|}
\hline
 & \multicolumn{2}{|c||}{\textbf{Avito}} & \multicolumn{2}{|c|}{\textbf{Company}} \\
\hline
\textbf{Algorithm} & AUC & Logloss & AUC & Logloss\\
\hline
LR & 0.7556 & 0.05918 & 0.7404 & 0.2404 \\
FM & 0.7802 & 0.06094 & 0.7557 & 0.2365 \\
DNN & 0.7816 & 0.05655 & 0.7579 & 0.2360 \\
PNN & 0.7817 & 0.05634 & 0.7593 & 0.2357 \\
Wide\&Deep & 0.7817 & 0.05595 & 0.7594 & 0.2355 \\
DeepFM & 0.7819 & 0.05611 & 0.7592 & 0.2358 \\
\hline
DeepCP & 0.7844 & 0.05546 & 0.7610 & 0.2354 \\
DeepMP & 0.7917 & 0.05526 & 0.7663 & 0.2345 \\
DeepMCP & \textbf{0.7927} & \textbf{0.05518} & \textbf{0.7674} & \textbf{0.2341} \\
\hline
\end{tabular}
\vskip -4pt
\caption{Test AUC and Logloss on two large-scale datasets. DNN = Pred, DeepCP = Pred+Corr, DeepMP = Pred+Match, DeepMCP = Pred+Match+Corr.}
\label{tab_auc}
\end{table}

\subsection{Evaluation Metrics} \label{sec_metric}
We use the following evaluation metrics.
\begin{enumerate}
\item \textbf{AUC}: the Area Under the ROC Curve over the test set. The larger the better. It reflects the probability that a model ranks a randomly chosen positive instance higher than a randomly chosen negative instance. A small improvement in AUC is likely to lead to a significant increase in online CTR \cite{cheng2016wide}.
\item \textbf{Logloss}: the value of Eq. (\ref{loss_p}) over the test set. The smaller the better.
\end{enumerate}

\subsection{Effectiveness}
Table \ref{tab_auc} lists the AUC and Logloss values of different methods. It is observed that FM performs much better than LR, because FM models second-order feature interactions while LR models linear feature importance. DNN further outperforms FM, because it can learn high-order nonlinear feature interactions \cite{he2017neural}. PNN outperforms DNN because it further introduces a production layer. Wide\&Deep further outperforms PNN, because it combines LR and DNN, which improves both the memorization and generalization abilities of the model. DeepFM combines FM and DNN. It performs slightly better than Wide\&Deep on the Avito dataset, but slightly worse on the Company dataset.

We now examine our proposed models. DeepCP contains the correlation subnet and the prediction subnet. DeepMP contains the matching subnet and the prediction subnet.
It is observed that both DeepCP and DeepMP outperform the best-performing baseline on the two datasets. As the baseline methods only consider the prediction task, these observations show that additionally consider representation learning tasks can aid the performance of CTR prediction. It is also observed that DeepMP performs much better than DeepCP. It indicates that the matching subnet brings more benefits than the correlation subnet. This makes sense because the matching subnet considers both the users and the ads, while the correlation subnet considers only the ads.
It is observed that DeepMCP that contains the matching, correlation and prediction subnets performs best on both datasets.
These observations demonstrate the effectiveness of DeepMCP.

\subsection{Effect of the Balancing Parameters}
\begin{figure}[!t]
\centering
\subfigure[Avito]{\includegraphics[width=0.23\textwidth, trim = 0 0 25 20, clip]{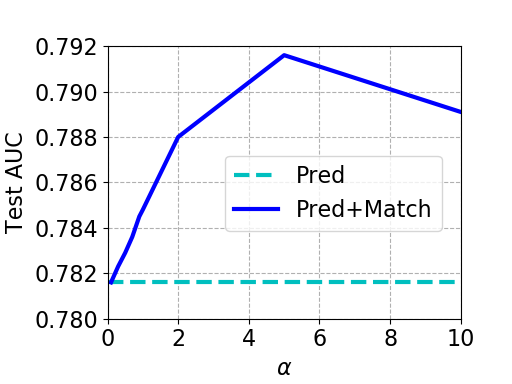}}
\subfigure[Company]{\includegraphics[width=0.23\textwidth, trim = 0 0 25 20, clip]{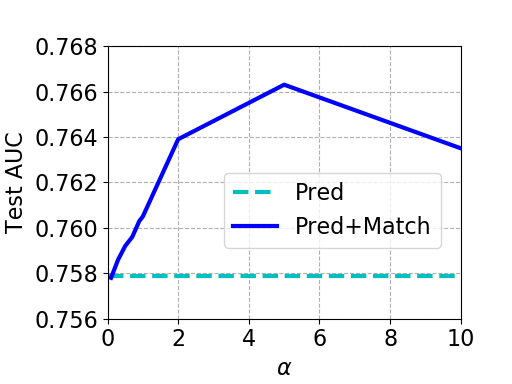}}
\vskip -8pt
\caption{AUC vs. $\alpha$ in DeepMCP ($\beta = 0$): effect of the matching subnet, in addition to the prediction subnet. DNN = Pred, DeepMP = Pred+Match.}
\vskip -10pt
\label{tune_m}
\end{figure}

\begin{figure}[!t]
\centering
\subfigure[Avito]{\includegraphics[width=0.23\textwidth, trim = 0 0 20 20, clip]{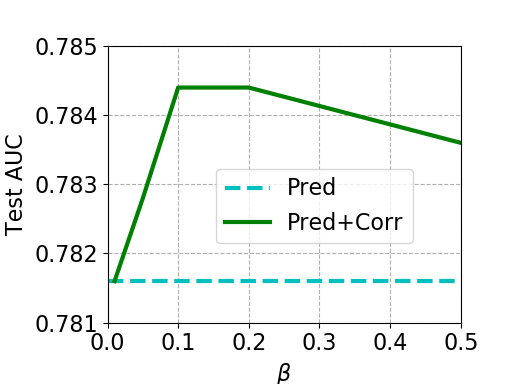}}
\subfigure[Company]{\includegraphics[width=0.23\textwidth, trim = 0 0 20 20, clip]{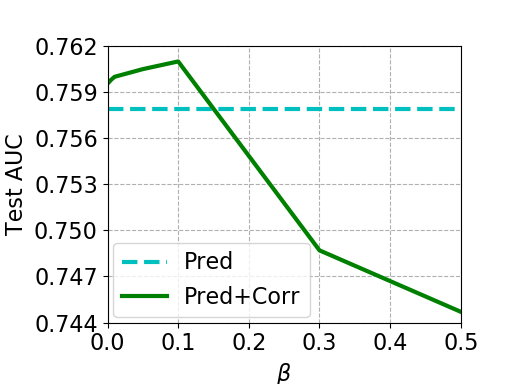}}
\vskip -8pt
\caption{AUC vs. $\beta$ in DeepMCP ($\alpha = 0$): effect of the correlation subnet, in addition to the prediction subnet. DNN = Pred, DeepCP = Pred+Corr.}
\vskip -10pt
\label{tune_c}
\end{figure}

In this section, we examine the effect of tuning balancing hyperparameters of DeepMCP.
Figure \ref{tune_m} and Figure \ref{tune_c} examine $\alpha$ (matching subnet) and $\beta$ (correlation subnet) respectively.
It is observed that the AUCs increase when one hyperparameter enlarges at the beginning, but then decrease when it further enlarges.
On the Company dataset, large $\beta$ can lead to very bad performance that is even worse than the prediction subnet only.
Overall, the matching subnet leads to larger AUC improvement than the correlation subnet.
The Company dataset is more sensitive to the $\beta$ parameter.

\subsection{Effect of the Hidden Layer Size}
\begin{figure}[!t]
\centering
\subfigure[Avito]{\includegraphics[width=0.23\textwidth, trim = 0 0 18 20, clip]{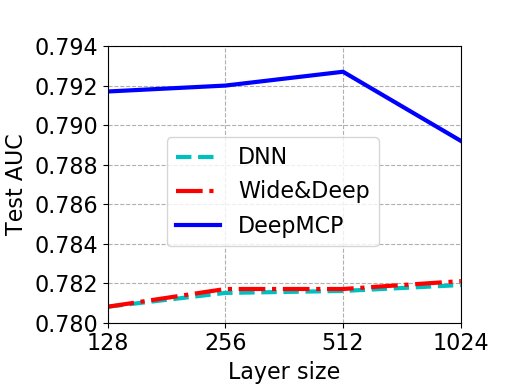}}
\subfigure[Company]{\includegraphics[width=0.23\textwidth, trim = 0 0 18 20, clip]{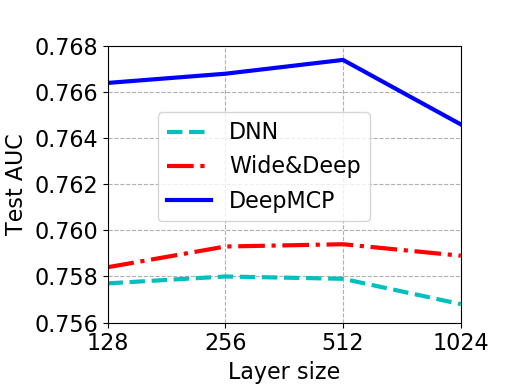}}
\vskip -8pt
\caption{AUC vs. hidden layer size. The number of hidden layers is 2. The x-axis is the dimension of the first hidden layer. The dimension settings are: [128, 64], [256, 128], [512, 256], [1024, 512].}
\vskip -10pt
\label{layer_size}
\end{figure}

In this section, we examine the effect of the hidden layer sizes of neural network-based methods.
In order not to make the figure cluttered, we only show the results of DNN, Wide\&Deep and DeepMCP. Figure \ref{layer_size} plots the AUCs vs. the hidden layer sizes when the number of hidden layers is 2. We use a shrinking structure, where the second layer dimension is only half of the first. It is observed that when the first layer dimension increases from 128 to 512, AUCs generally increase. But when the dimension further enlarges, the performance may degrade. This is possibly because it is more difficult to train a more complex
model.

\subsection{Effect of the Number of Hidden Layers}
\begin{figure}[!t]
\centering
\subfigure[Avito]{\includegraphics[width=0.23\textwidth, trim = 0 0 25 20, clip]{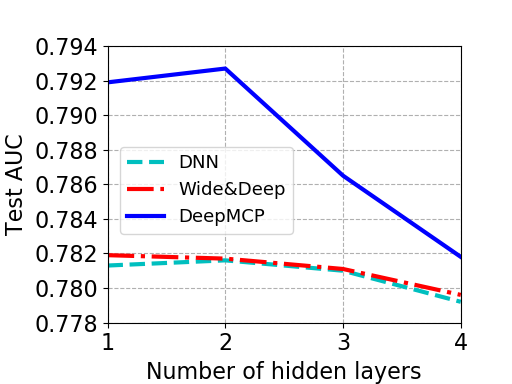}}
\subfigure[Company]{\includegraphics[width=0.23\textwidth, trim = 0 0 25 20, clip]{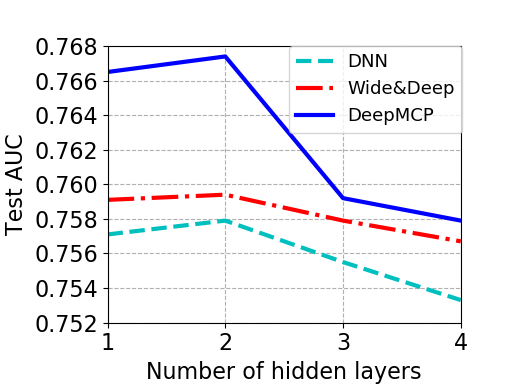}}
\vskip -8pt
\caption{AUC vs. the number of hidden layers. The dimension settings are: 1 layer - [256], 2 layers - [512, 256], 3 layers - [1024, 512, 256], and 4 layers - [2048, 1024, 512, 256].}
\vskip -10pt
\label{layer_depth}
\end{figure}

In this section, we examine the effect of the number of hidden layers.
The dimension settings are: 1 layer - [256], 2 layers - [512, 256], 3 layers - [1024, 512, 256], and 4 layers - [2048, 1024, 512, 256]. It is observed in Figure \ref{layer_depth} that when the number of hidden layers increases from 1 to 2, the performance generally increases. This is because more hidden layers have better expressive abilities \cite{he2017neural}. But when the number of hidden layers further increases, the performance then decreases. This is because it is more difficult to train deeper neural networks.

\section{Related Work}
\subsubsection{CTR Prediction}
CTR prediction has attracted lots of attention from both academia and industry \cite{he2014practical,cheng2016wide,he2017neural,zhou2018deep}.
Generalized linear models, such as Logistic Regression (LR) \cite{richardson2007predicting} and Follow-The-Regularized-Leader (FTRL) \cite{mcmahan2013ad}, have shown decent performance in practice. However, a linear model lacks the ability to learn sophisticated feature interactions \cite{chapelle2015simple}. Factorization Machines (FMs) \cite{rendle2010factorization,rendle2012factorization} are proposed to model pairwise feature interactions in terms of the latent vectors corresponding to the involved features. Field-aware FM \cite{juan2016field} and Field-weighted FM \cite{pan2018field} further consider the impact of the field that a feature belongs to in order to improve the performance of FM.

In recent years, Deep Neural Networks (DNNs) are exploited for CTR prediction and item recommendation in order to automatically learn feature representations and high-order feature interactions \cite{van2013deep,covington2016deep,wang2017deep,he2017neural}.
Zhang et al. [\citeyear{zhang2016deep}] propose Factorization-machine supported Neural Network (FNN), which pre-trains an FM before applying a DNN. Qu et al. [\citeyear{qu2016product}] propose the Product-based Neural Network (PNN) where a product layer is introduced between the embedding layer and the fully connected layer. Cheng et al. [\citeyear{cheng2016wide}] propose Wide\&Deep, which combines LR and DNN in order to improve both the memorization and generalization abilities of the model.
Guo et al. [\citeyear{guo2017deepfm}] propose DeepFM, which models low-order feature interactions like FM and models high-order feature interactions like DNN.
He et al. [\citeyear{he2017neural}] propose the Neural Factorization Machine which combines the linearity of FM and the non-linearity of neural networks.
Nevertheless, these methods mainly model the feature-CTR relationship. Our proposed DeepMCP model further considers user-ad and ad-ad relationships.

\subsubsection{Multi-modal / Multi-task Learning}
Our work is also closely related to multi-modal / multi-task learning, where multiple kinds of information or auxiliary tasks are introduced to help improve the performance of the main task. For example, Zhang et al. [\citeyear{zhang2016collaborative}] leverage heterogeneous information (i.e., structural content, textual content and visual content) in a knowledge base to improve the quality of recommender systems. Gao et al. [\citeyear{gao2018recommendation}] utilize textual content and social tag information, in addition to classical item structure information, for improved recommendation.
Huang et al. [\citeyear{huang2018improving}] introduce context-aware ranking as an auxiliary task in order to better model the semantics of queries in entity recommendation.
Gong et al. [\citeyear{gong2019deep}] propose a multi-task model which additionally learns segment tagging and named entity tagging for slot filling in online shopping assistant. In our work, we address a different problem and we introduce two auxiliary but related tasks (i.e., matching and correlation with shared embeddings) to improve the performance of CTR prediction.

\section{Conclusion}
In this paper, we propose DeepMCP, which contains a matching subnet, a correlation subnet and a prediction subnet for CTR prediction.
These subnets model the user-ad, ad-ad and feature-CTR relationship respectively. Compared with classical CTR prediction models that mainly consider the feature-CTR relationship, DeepMCP has better prediction power and representation ability. Experimental results on two large-scale datasets demonstrate the effectiveness of DeepMCP in CTR prediction.
It is observed that the matching subnet leads to higher performance improvement than the correlation subnet.
This is possibly because the former considers both users and ads, while the latter considers ads only.

\balance
\bibliographystyle{named}
\bibliography{ref}

\end{document}